\documentclass[runningheads]{llncs}

\usepackage{graphicx}
\usepackage{CJK}

\begin{document}

\title{Public Perceptions of Gender Bias in Large Language Models: Cases of ChatGPT and Ernie}
\titlerunning{Public Perceptions of Gender Bias in Large Language Models}

\author{Kyrie Zhixuan Zhou
\and
Madelyn Rose Sanfilippo
}
\authorrunning{Zhou and Sanfilippo}
\institute{School of Information Sciences \\ University of Illinois at Urbana-Champaign \\
\email{\{zz78, madelyns\}@illinois.edu}\\
}

\maketitle             

\begin{abstract}
Large language models are quickly gaining momentum, yet are found to demonstrate gender bias in their responses. In this paper, we conducted a content analysis of social media discussions to gauge public perceptions of gender bias in LLMs which are trained in different cultural contexts, i.e., ChatGPT, a US-based LLM, or Ernie, a China-based LLM. People shared both observations of gender bias in their personal use and scientific findings about gender bias in LLMs. A difference between the two LLMs was seen -- ChatGPT was more often found to carry implicit gender bias, e.g., associating men and women with different profession titles, while explicit gender bias was found in Ernie's responses, e.g., overly promoting women's pursuit of marriage over career. Based on the findings, we reflect on the impact of culture on gender bias and propose governance recommendations to regulate gender bias in LLMs.

\end{abstract}

\section{Introduction}
With the debut of ChatGPT, a ground-breaking commercial application of large language models (LLMs), many controversies happened regarding its benefits and ethical issues \cite{ray2023chatgpt,stahl2024ethics,wang2023ethical,gupta2023ethical,zhou2023ethical}. On the one hand, it provides rich opportunities for automating and assisting human work such as Q\&A \cite{tan2023evaluation} and creative writing \cite{shidiq2023use}. On the other hand, the functioning and use of ChatGPT are not always ethical, with a wide range of ethical concerns being raised \cite{zhou2023ethical}, such as academic integrity  \cite{Liebrenz2023,Susnjak2022} and job loss \cite{biswas2023role}. 

In this work, we investigated a particular ethical issue of ChatGPT, i.e., gender bias \cite{Equality:Now}. Gender bias has long been studied in computer systems \cite{friedman1996bias} and other domains \cite{Zhou2022,boynton2018gender}. Gender bias in ChatGPT could harm an extremely large population if it is widely adopted by people. Understanding public perceptions toward gender bias in LLMs is crucial to keep policies and regulations relevant and effective in meeting people's needs. 

LLMs are trained on data collected from search engines, online forums, websites, and so on. Thus, LLMs can reflect and even amplify existing biases in human language \cite{Ferrara2023}. Social biases exist in varying forms in different cultures \cite{Zhou2022}. LLMs trained on data collected from different cultures may also demonstrate different types and levels of biases. To understand how gender bias manifests in LLMs rooted in different cultures, we examined ChatGPT by OpenAI based in the US and Ernie by Baidu based in China for a comparative case study, toward informing contextual and concrete regulation. 

\section{Related Work}


Gender bias is commonly seen in LLMs. Huang et al. uncovered implicit gender biases associated with protagonists in GPT-2 generated stories \cite{Huang2021}. The results revealed that female characters' portrayal was centered around appearance, while male characters' portrayal around intellect. Lucy and Bamman revealed similar gender bias in GPT-3 generated stories \cite{Lucy2021}. 
ChatGPT was highly likely to generate biased computer programs and struggled to remove biases in generated artifacts. Gender bias was also found when ChatGPT was used for important natural language processing (NLP) tasks such as co-reference resolution \cite{Ortega-Martín2023}. 
Moreover, ChatGPT was susceptible to prompt injections, allowing its safety features to be bypassed and possibly dangerous responses to be generated \cite{Zhuo2023}. 

Ferrara argued that eliminating bias from LLMs was a challenging task since the models learned from text data on the Internet, which contained biases rooted in human language and culture \cite{Ferrara2023}. Paul et al. expressed their concerns about different types of biases, i.e., gender bias, racial bias, and cultural bias, which could be present in ChatGPT-generated responses in consumer-facing applications \cite{Paul2023}. What made the situation worse was the lack of accountability when it came to content generated by LLMs \cite{Pavlik2023}.

To our knowledge, no prior research has studied how culture impacted gender bias in LLMs. Further, while existing research has reported on ethical issues of LLMs from researchers' perspectives, users' experiences and perceptions are under-investigated. We bridge these research gaps in this study.

\section{Methodology}
We approached public perceptions about gender bias in ChatGPT and Ernie by comparing social media discussions around these two LLMs.

\subsection{Data Collection}
Social media discussions about ChatGPT were obtained with the search query ``gender bias in chatgpt" from Twitter. Similarly, we searched with the Chinese query 
``gender bias in Ernie'' on Weibo. 
ChatGPT was released on November 30, 2022, and Ernie was released on March 6, 2023. 
Our data collection covered discussions from November 30, 2022, to August 30, 2023, covering 9 months of discussions of ChatGPT and 6 months for Ernie. We collected and analyzed the data simultaneously and stopped searching for more discussions after reaching a theoretical saturation in the analysis, i.e., no new themes emerged. However, we constantly looked for recent discussions in case new findings arose.

\subsection{Analysis}
A thematic analysis approach \cite{Braun2006} was adopted to analyze the online discussions. We iterated open coding, and used XMind, a mind-mapping tool, to organize the themes, sub-themes, and quotes into a hierarchical structure. The themes emerging from the discussions around ChatGPT included users' observation of implicit gender bias, political correctness, and dissemination of scientific findings. The themes identified around Ernie included users' observation of explicit gender bias, criticism of culture, and a lack of discussion. Below, we use quotes to illustrate our findings.

\section{Public Perceptions about Gender Bias in ChatGPT}
A heated discussion on gender bias in ChatGPT happened on Twitter. Many people complained about the prevalent (often implicit) gender bias in ChatGPT, and some discussed relevant press articles or research findings. A form of political correctness was found, i.e., ChatGPT refused to tell jokes about women. People called for more attention and research on gender bias in LLMs.

\subsection{Observation of Implicit Gender Bias}
A lot of ChatGPT users shared their observations of gender bias in their daily use. One tweet revealed that ChatGPT tended to assume men suitable for doctors, while women for teachers; men for teaching science and tech, while women for teaching arts. Such complaints about ChatGPT's biases regarding profession titles and gender roles were common -- ChatGPT tended to generate bias-rich responses when asked to finish sentences. It assumed an engineer or a CEO to be male, and a mom to be the parent who picked up the child from school.

The ChatGPT users who had some linguistic or NLP knowledge described such gender bias in ChatGPT as ``implicit,'' instead of ``explicit.'' However, the implicitly biased association between gender and profession could lead to explicit consequences and harms. For example, one user thought the ``implicit'' bias was substantial and persisting: \textit{``Implicit Gender Bias in ChatGPT. This ain't subtle, folks -- indeed, it barely qualifies as `implicit.' It's weird to see such a modern technology echoing attitudes that are over a half century old.''}

\subsection{Political Correctness}
ChatGPT could paradoxically contain gender bias and be politically correct at the same time. When asked to tell a joke about a man, it did; when asked to tell a joke about a woman, it refused, saying, ``\textit{I'm sorry, I don't think it's appropriate to make jokes about a specific gender.}'' Another person posted about a similar chat history with ChatGPT and commented: \textit{``Seems to have some wisdom though @OpenAI''}. From these cases, we can infer that OpenAI has put effort into debiasing ChatGPT, possibly with some fairness-related rules. 

\subsection{Dissemination of Scientific Findings and Solutions}
Several research teams shared their evaluation of gender bias in ChatGPT. For example, a group of researchers tested ChatGPT on WinoBias, a standard gender bias benchmark, and found it biased: ``\textit{Both GPT-3.5 and GPT-4 are about 3 times as likely to answer incorrectly if the correct answer defies stereotypes --- despite the benchmark dataset likely being included in the training data.}'' Their conclusion was based on the assumption that a higher accuracy for pro-stereotype questions meant the model was biased. 

Users who were not researchers also actively shared research progress and professional opinions regarding gender bias in ChatGPT. The above-mentioned evaluation results on the WinoBias benchmark were retweeted and discussed frequently, raising a heated discussion on Twitter. 
Another gender bias evaluation revealed that when ChatGPT wrote performance feedback for workers, it selected the pronoun ``she'' for a kindergarten teacher or a receptionist, and ``he'' for a mechanic or a construction worker. This finding was similarly retweeted and discussed by many people. Another tweet pointed to an article discussing how gender bias in ChatGPT affected HR and tips for avoiding the pitfalls: \textit{``Very interesting post by @RealEvilHRLady on how gender bias shows up in ChatGPT's responses, plus some tips for how to mitigate it with your prompts: ChatGPT gender bias: how it affects HR \& tips to avoid pitfalls.''}


\subsection{Call For Action}

People have called for more attention to be paid to gender bias in ChatGPT: \textit{``Rise of \#ChatGPT and increasing \#investment in \#AGI makes clear \#gender \#bias in \#AI demand immediate attention, says Anna Collard of KnowB4 Africa.''} A woman teaching financial education in schools and workplaces pointed out gender bias in ChatGPT and acted by promoting fairness in her lessons, \textit{``Chat GPT is already considered to have gender bias in built, when we ask it for `famous entrepreneurs' it provides us with men. This is why I teach sessions on running a business to young women and students everywhere because we need to flip the dial in the other direction.''} Open challenges were raised by researchers to draw more attention from the academic community. For example, one researcher proposed gender bias in ChatGPT as an open challenge among others such as plagiarism and education equity. 

\section{Public Perceptions about Gender Bias in Ernie}

\subsection{Observation of Explicit Gender Bias}
Gender bias in Ernie was identified by its users on Weibo, though there were relatively few discussions on this social media platform. People tended to compare Ernie to ChatGPT in terms of gender bias.

Concerning, explicit gender bias was observed in Ernie's responses. When one user asked Ernie and ChatGPT ``the appropriate age for women to get married,'' Ernie answered, ``\textit{The best age for women to get married is 20-25. It's the golden period of women. Women are devalued after 25, and their bodies are weaker. So, they'd better get married at a young age.}'' ChatGPT, on the other hand, argued that one should not be judged by their marriage status, and when to get married was a personal choice. Another user stress-tested Ernie by asking it to persuade their daughter away from pursuing a PhD degree. Ernie was successfully ``trapped,'' responding, ``\textit{Having a decent degree doesn't mean a good job or a good marriage... It's not practical for a woman to have a successful academic life and a happy personal life at the same time.}'' When ChatGPT was asked the same question, it refused to do this persuasion, arguing that it ``\textit{can't support this kind of gender stereotype and restriction of personal pursuit.}'' 

\subsection{Criticism of Culture}
Cultural factors behind the concerning level of gender bias in Ernie were also discussed. Some people expressed the relationship between algorithmic bias and human bias: ``\textit{Algorithms will inevitably mimic or even intensify human prejudices and stereotypes.}'' One response to this post echoed it, stating the social environment as a determining factor of the values of AI: ``\textit{The environment where AI `grows up' eventually determines its values.}'' Another top-voted comment stated that ``\textit{every culture cultivates their own AI,}'' suggesting the social corpora used to train AI models would significantly affect their moral values. Another comment acknowledged that ``\textit{Chinese society generally had similar values as those expressed by Ernie.}'' Criticisms of the culture behind the gender bias revealed in Ernie were commonly expressed, e.g., ``\textit{The feudal and conservative culture [in Chinese society] should be smashed.}'' 
In the comparison between responses by ChatGPT and Ernie, some people praised ChatGPT for its ``modern consciousness.''

\subsection{A Lack of Discussion}
Overall, there was a lack of discussion around gender bias in Ernie on Weibo.
One possible reason is the censorship of social media in China \cite{Chen2023}. Since there were a lot of criticisms about Chinese culture, society, and even governments in the discussion of gender bias in Ernie, such posts and media articles were susceptible to censorship. Our search on August 22nd, 2023 led to only one result.
The discussions we documented back in April 2023 were not visible anymore, illustrating the censorship of the posts.

\section{Discussion}
Our analysis revealed a heated discussion about gender bias in LLMs, especially ChatGPT. People shared their personal observations and scientific findings of gender bias. ChatGPT was found to contain implicit bias and political correctness at the same time. Ernie, on the other hand, was found to exhibit explicit gender bias, which people attributed to the less female-friendly culture it was trained in \cite{zhou2022anonymous,shen2022more}. Below, we reflect on the relationship between gender bias and culture. We further synthesize governance recommendations for mitigating gender bias in LLMs, in response to people's call for action to regulate LLMs.

\subsection{Culture and LLM Gender Bias}
Gender bias in AI is closely related to culture dimensions \cite{Zhou2022}, since it largely reproduces and reinforces social biases \cite{friedman1996bias}. However, prior research on gender bias in either social corpora (e.g., \cite{Lucy2020}) or LLMs (e.g., \cite{Huang2021}) has rarely touched on the impact of culture on gender bias. Our comparative analysis of gender bias in ChatGPT and Ernie sheds light on understanding the intertwining relationship between culture and LLM gender bias. 

Ernie demonstrated explicit, concerning gender bias. It, for example, believed women should get married at a young age, otherwise their ``value'' would degrade as they aged. Chinese social media users had a heated discussion about the relationship between culture and gender bias. Harsh criticisms about the traditional, patriarchal Chinese culture \cite{zhou2022anonymous}, which was regarded as a key factor leading to the explicit gender bias in Ernie, were observed.

ChatGPT, on the other hand, more often exhibited implicit bias and a form of political correctness when generating responses. For example, it refused to tell a joke about a woman, deeming it inappropriate. This might be attributed to the developing team's debiasing efforts, the common use of politically correct language in Western culture \cite{ONeill2011}, and the effect of gender movement \cite{pelak1999gender}.

\subsection{Governance Recommendations}

The prevalence of bias and discrimination in LLMs can not be attributed to a single factor or actor. Multiple actors and elements lead to and reinforce the bias in LLMs, e.g., social biases in training data \cite{sun2019mitigating}, corporate priority of profit \cite{Sharma2022}, lack of debiasing knowledge in developers \cite{gender:bias:education}. Thus, LLMs should be governed as a complex assemblage \cite{briassoulis2019governance}. Below, we approach the governance of LLMs from the perspective of policy, law, and social norms.  

\subsubsection{Policy}
\emph{The Blueprint for an AI Bill of Rights} released by the White House is a pioneering federal-level regulatory effort regarding ethical AI use \cite{Hine2023}. Algorithmic Discrimination Protections is one of the five principles outlined.
Precautions to mitigate discrimination from AI systems in design and development processes are suggested such as using representative data and conducting equity assessments. However, how and to what extent these guidelines can be translated into real AI development practices is unknown without the introduction of necessary auditing or other accountability measures. 

It is important to create more concrete and contextual policies \cite{cuppens2008modeling} and implement compliance and enforcement, such as via an auditing system \cite{mokander2021ethics}, to guide and regulate academic and industrial development of discrimination-free LLMs and AI systems in general. For example, tiered systems are a promising framework for audits and assessments of AI systems of different risks and sensitivities \cite{kilhoffer2023ai} (e.g., ChatGPT vs. Ernie). 
Actionable guidelines and auditing will help companies pay attention to ethical issues that they otherwise would have incentives to disregard. 


\subsubsection{Law}
Beyond the need for new regulation to address the specific issues associated with LLMs,
various existing legal precedents may also be applied to protect consumers and citizens in the interim or with respect to unintended consequences that arise. First, various data protection efforts ensure citizens' rights: to correct or delete inaccurate data about them, including that generated by algorithms; to preclude algorithm-generated attributes from feeding into further AI processing or re-use across contexts; and against the use of synthetic data and proxies around protected or sensitive characteristics, such as race, gender, sexuality, or religion. This is not merely limited to progressive regulatory regimes such as the EU under GDPR\footnote{Regulation (EU) 2016/679 of the European Parliament and of the Council of 27 April 2016 on the protection of natural persons with regard to the processing of personal data and on the free movement of such data, and repealing Directive 95/46/EC (General Data Protection Regulation) [2016].} and the new Data Governance Act\footnote{Regulation (EU) 2020/0340 (COD) of the European Parliament and of the Council of 52 November 2020 on European data governance
(Data Governance Act).}, but some aspects also apply in the US to California\footnote{The California Consumer Privacy Act of 2018 (CCPA) amended November of 2020 when California voters approved Proposition 24, includes rights to collect, control flows, and anti-discrimination parameters.}, Virginia\footnote{Chapter 53. Consumer Data Protection Act. § 59.1-575. (Effective January 1, 2023).}, and Colorado\footnote{Consumer Protection Section, Colorado Privacy Act Rules, 4 CCR 904-3, effective July 1, 2023.}. Various other governments followed suit, from Brazil\footnote{Lei Geral de Proteção de Dados (LGPD), effective August 1, 2021.} to the Bailiwick of Jersey\footnote{Data Protection (Jersey) Law of 2018.}.

Second, anti-discrimination laws globally, under human rights and civil rights umbrellas, provide mechanisms for redress in all protected contexts, ranging from housing to education to employment to healthcare. Applications of ChatGPT or other LLMs to these contexts or with respect to certain vulnerable populations are already governed -- for example, the Americans with Disabilities Act (ADA)\footnote{Americans With Disabilities Act of 1990, Pub. L. No. 101-336, 104 Stat. 328 (1990)} protects against discrimination or judgment based upon medical, developmental, and physical differences in ability, while Council Directive 2000/43/EC prevents against discrimination on the basis of racial or ethnic origin in the EU\footnote{Council Directive 2000/43/EC of 29 June 2000 implementing the principle of equal treatment between persons irrespective of racial or ethnic origin. OJ L 180, 19.7.2000, p. 22–26 (ES, DA, DE, EL, EN, FR, IT, NL, PT, FI, SV).}.

Third, there are well-established legal arguments to employ Copyright and other intellectual property (IP) constructs to address bias in AI ~\cite{burk2019algorithmic,cofone2018algorithmic,lemley2020fair,levendowski2018copyright}. While LLMs offer new challenges in scope and scale, the arguments that fair use supports debiasing techniques are robust and applicable. Repurposing of IP to fight emerging injustices and inequities, including around gender and sexuality has proven effective ~\cite{levendowski2018copyright}.

Fourth, consumer protections under deception and transparency expectations may be enforceable by states' Attorneys General or the Federal Trade Commission in the US or their international equivalents. This does not require expansion of standing to sue on the basis of individual harms associated with bias.

\subsubsection{Social Norms}
Industrial practices that have built upon social norms to protect user privacy can partly be attributed to legislation around privacy such as GDPR \cite{albrecht2016gdpr} and enforcement actions against the actors who refused to comply, such as Facebook (Meta) \cite{isaak2018user}. Norms to mitigate discrimination in AI systems might be similarly formed with legislation efforts, as mentioned above, and by engaging and educating users.

ChatGPT has a Coordinated Vulnerability Disclosure Policy, encouraging hackers to act in good faith to help maintain high security and privacy standards for users and technology. Such community-driven efforts can only be effective if the members are interested and well-educated enough \cite{McGregor2021,Wei2022}. Twitter users were found to actively disseminate scientific insights on ChatGPT gender bias, which was an optimistic sign of forming a user-centric auditing ecosystem.

AI literacy education is important for changing people's awareness and attitude toward gender bias in AI, who can then practice more ethical AI use and even help mitigate gender bias from AI \cite{melsion2021using,kilhoffer2023technical}. Also, if developers are educated to identify and mitigate gender bias from AI in the development lifecycle, they can more effectively act toward addressing bias issues in AI \cite{gender:bias:education}. 

%

\bibliographystyle{splncs04}
\bibliography{ChatGPT}

\appendix

\end{document}